# Constructing Belief Networks to Evaluate Plans

Paul E. Lehner*
Systems Engineering. Department
George Mason University
Fairfax, VA 22030

plehner@mason1.gmu.edu

Christopher Elsaesser
AI Technical Center
The MITRE Corporation
7525 Colshire Drive
McLean, VA 22102
chris@starbase.mitre.org

Scott A. Musman
AI Technical Center
The MITRE Corporation
7525 Colshire Drive
McLean, VA 22102
musman@starbase.mitre.org

## Abstract

This paper examines the problem of constructing belief networks to evaluate plans produced by an knowledge-based planner. Techniques are presented for handling various types of complicating plan features. These include plans with context-dependent consequences, indirect consequences, actions with preconditions that must be true during the execution of an action, contingencies, multiple levels of abstraction, multiple execution agents with partially-ordered and temporally overlapping actions, and plans which reference specific times and time durations.

## 1. INTRODUCTION

Uncertainty is ubiquitous in planning problems. Despite this, few knowledge-based planning systems have been developed that can reason *explicitly* about uncertainty. Instead, most knowledge-based planning systems are based solely on symbolic reasoning (Allen, et. al., 1990). Although these systems may employ techniques that adapt a plan to unanticipated events, they cannot generate quantitative uncertainty estimate of possible future states.

Consequently, the best these planners can do is react. They cannot generate plans that are deduced to be robust against probable futures.

Recently a number of researchers have recognized the importance of uncertainty in automated planning and are developing approaches to address it (e.g., Dean & Wellman, 1991; Hanks, 1990; Kushmerick, et.al, 1993). Common to many of these approaches is the use of *belief networks* to represent and reason about uncertainties in plans. To date, however, research in the use of belief networks to reason about uncertainty in planning has been restricted to limited types of plans. Most, for instance, assume a single execution agent, a single level of abstraction and no contingencies.

If belief networks are to provide a foundation for probabilistic planning, then we need to examine the extent to which different plan features can be represented in belief networks. This paper examines this issue. In particular we overview how to develop belief networks that can handle a variety of plan features. All of the capabilities described below are being implemented as part of the AP planning system. The AP system is designed for adversarial planning problems where each planning agent may have multiple execution agents that execute coordinated activities (Elsaesser and MacMillan, 1991).

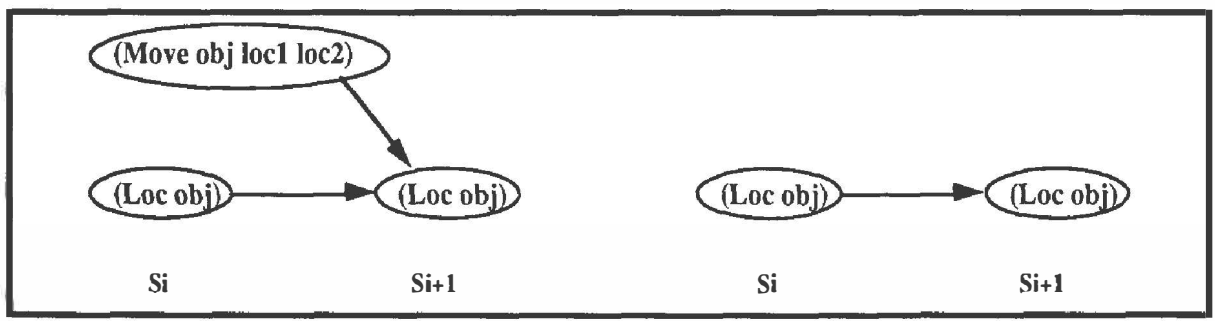

**Figure 1. Example of Action and Persistence Models.**

---

## 2. BASIC APPROACH

The basic idea behind using belief networks for plan evaluation is to construct a belief network from a knowledge base of *probabilistic action models* and *probabilistic persistence models*. (Wellman, 1990). A probabilistic action model specifies probability distributions on a set of *consequence predicates* conditioned on the state of a set of *predecessor predicates*. For example, the probability action model depicted in Figure 1 asserts that the location of the object referenced by *obj* ( (Loc obj) ) in the situation after the action of moving *obj* from *loc1* to *loc2* ( (Move obj loc1 loc2) ) is completed (situation $S_{i+1}$) is a probabilistic function of the location of *obj* in the prior situation. Similarly, the probabilistic persistence model for (Loc obj) is a probabilistic function of the state of (Loc obj) in the previous state.

Consider the two step plan (Move A L1 L2) --> (Move B L3 L1). To build a belief network to evaluate this plan, one can begin by sequentially *pasting onto* the belief network the probabilistic action model for each action (Figure 2a). When pasting onto the belief network, conflicting information already in the network is replaced. Note that the network in Figure 2a is incomplete, since there are nodes in future states which are not connected to the current state. To complete this network, it is necessary to work backwards through the network and sequentially *pasting into* the network the necessary persistence models (Figure 2b). When pasting into the network, current entries in the network are not changed, but previously unspecified nodes and probability assessments may be entered.

We refer to a belief network, such as shown in Figure 2b, as a *plan evaluation network* or PE-net. Once constructed, existing algorithms can be applied to the PE-net to calculate the marginal probability of any node in the PE-net as a function of information about the initial or future state.

## 3. PLAN FEATURES THAT COMPLICATE PE-NET CONSTRUCTION.

PE-net construction is straight forward for simple linear plans such as the one mentioned above. However, as plans get more complex, the process of constructing a PE-net becomes correspondingly more complex. Below we show how to handle a number of these complexities.

### 3.1 PARTIAL MODELS

The PE-net approach to plan evaluation assumes a knowledge base of action and persistence models, each of which is a small, paritially-specified belief network. Given the number of actions and predicates that may be mentioned in the knowledge base, it is unlikely that all of the conditional probabilities mentioned in all of these networks will be specified. It is more likely that the probabilities for consequence predicates will only be specified for a subset of the predecessor states. We handle this as follows. Wherever the action model is under specified, we paste into the PE-net the persistence models for the consequent predicates. Wherever the persistence model is under specified, we paste into the PE-net a default persistence model. For our applications the default persistence model asserts that no change will take place.

Using this technique all the conditional probabilities in the network will be specified. All that remains is to specify the unconditional probabilities for the initial state.

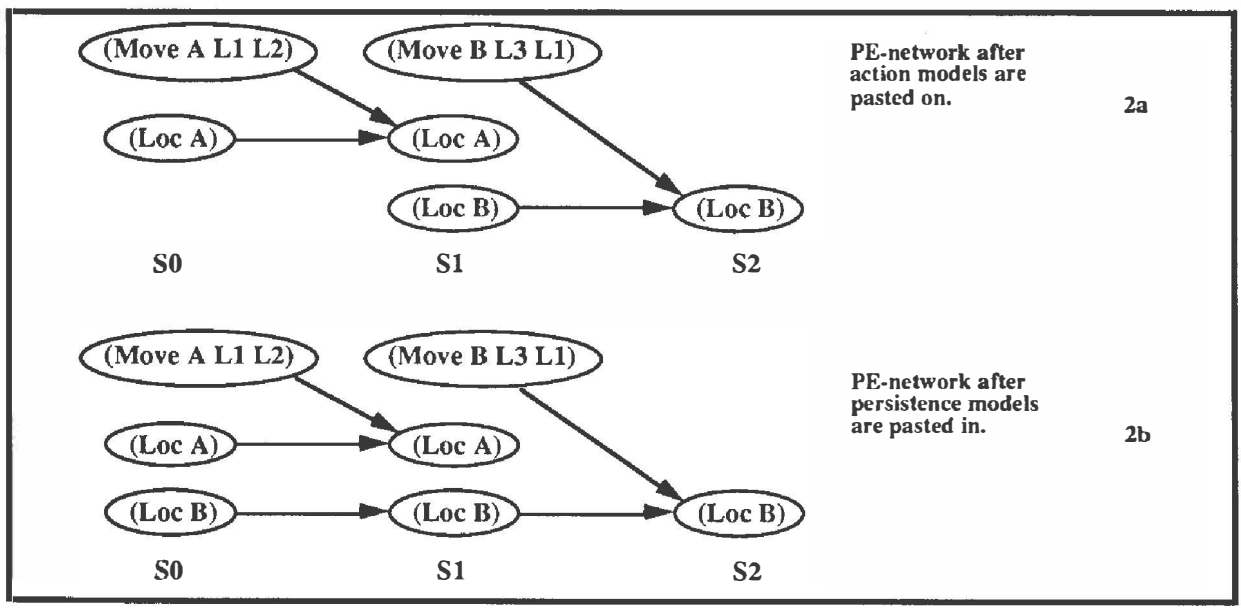

**Figure 2.  Constructing a PE-net.**

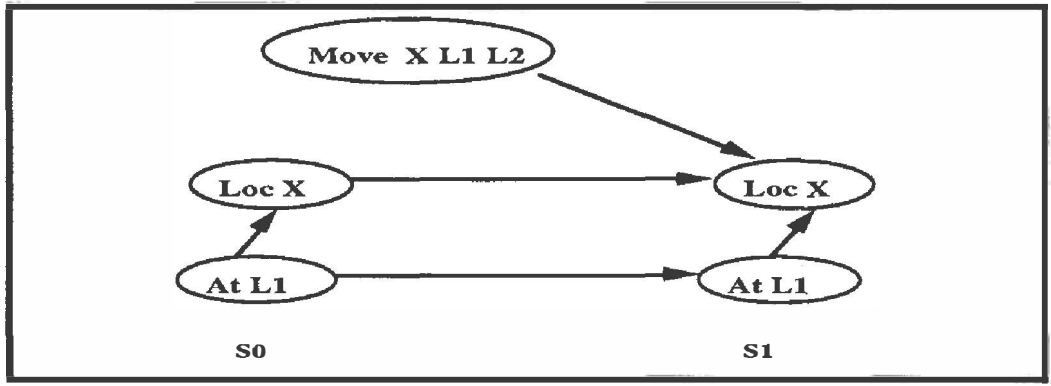

**Figure 3.    Problematic PE-net with derived effects.**

## 3.2 DERIVED EFFECTS

In developing a PE-net, it is important to separate *causal effects* from *derived effects*. Causal effects are links that go from a predicate node in one situation to a predicate node in a later situation. Derived effects are defined by links between two predicate nodes in the same situation.

To illustrate the kind of problem that may be encountered, consider the simple PE-net in Figure 3. This PE-net is for a single Move action. It also includes (At L1) nodes which indicate what object is at location L1. Clearly, if (At L1)=X then (Loc X)=L1. Consequently, the status of (Loc X) can sometimes be derived from the status of (AT L1) in the same situation. It seems natural therefore to paste onto the PE-net an arc from (At L1) to (Loc X), where the conditional probability P((Loc X)=L1|(At L1)=X & anything else)=1 is specified. This is an example of a derived effect. Now, assume that the move action is completely reliable, all persistence models are the no change default model, and X is initially at L1.

Given these assumptions, we would expect (Loc X)=L2 in $S_1$ with certainty. However, the PE-net in Figure 3 implies (Loc X)=L1 in $S_1$ with the certainty! The addition of the derived effect unexpectedly resulted in the persistence model for (AT L1) overriding the action model.

In general, this type of problem occurs because derived effects serve to complete an incomplete causal model. In theory, it is possible to do away with derived effects altogether. If causality is temporal, then a complete causal model going from $S_i$ to $S_{i+1}$ would account for all interactions within a situation. In Figure 3, for instance, a complete action model would have both (Loc X) and (At L1) as consequence predicates. This would remove the need to directly connect (Loc X) and (At L1) in $S_1$. Unfortunately, the knowledge engineering effort required to develop a complete causal model is prohibitive, since it would require the specification of conditional probabilities for all direct and indirect consequences of an action.

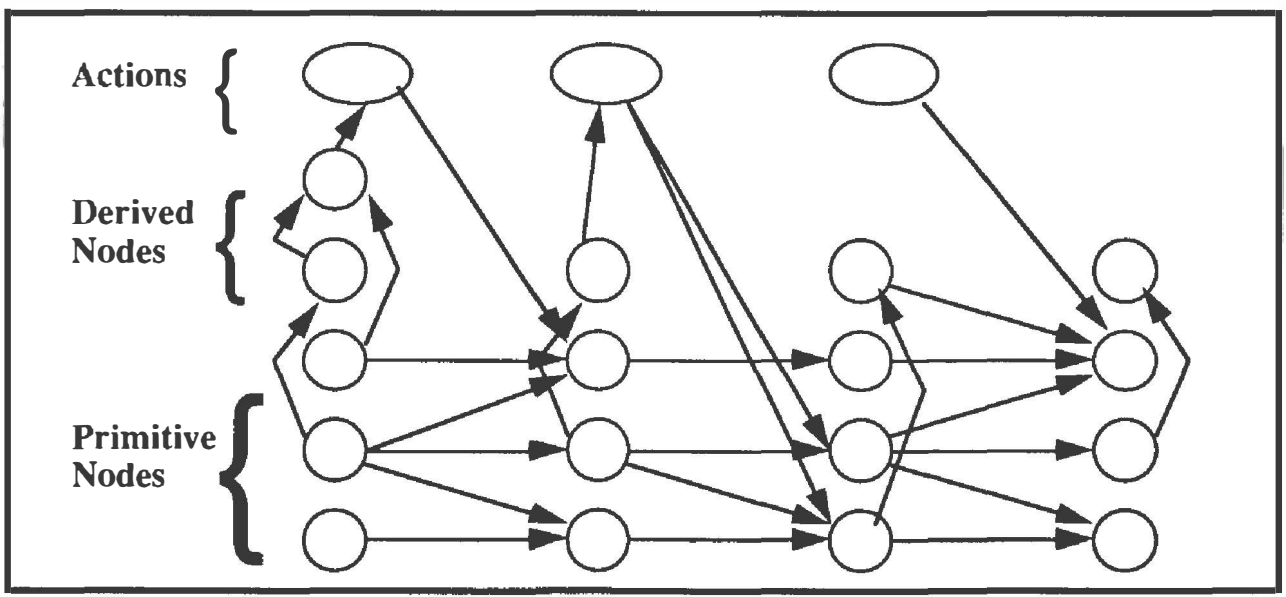

**Figure 4.   PE-net with primitive and derived predicates**

Our approach to derived effects is a compromise between complete causal modeling and the liberal use of derived effects. All the PE-nets constructed by our system generates networks with the structure depicted in Figure 4. Predicates are split into two levels. *Primitive* predicates do not have interconnections within a situation. It is assumed that they are only conditioned on the state of the nodes in the previous situation. It is up to the knowledge engineer of the action and persistence models to ensure that the models are causally complete with respect to primitive predicates. Predicates at the *derived level* can only be conditioned on other nodes in the same situation. The predicates at the derived level change from situation to situation. Only relevant derived-level predicates are included. Enforcing this structure removes the problems with derived effects.

This approach requires that any predicate mentioned as a consequence in an action model must be a primitive predicate. In Figure 3, therefore, (Loc X) would need to be a primitive node, (At L1) a derived node, and the arcs would go from (Loc X) to (At L1). This chnage would repair the problem in Figure 3.

### 3.3 CONTEXT-DEPENDENT EFFECTS

Many planners have actions models where the consequences of an action are functions of the situation in which the action was executed (Wilkins, 1988). In a PE-net this can be handled by invoking these same functions to determine possible node states in situation $S_{i+1}$ as a function of the possible node states in situation $S_i$. Iterating through the states in this way will enumerate all possible states for each node.

### 3.4 PLANNED CONTINGENCIES

A plan contains contingent actions when the decision to execute an action (or which action) is contingent on the situation. In a PE-net, contingent actions can be handled by combining actions into a single node, and then conditioning the merged action node on the nodes which determine which action will be executed.    Figure 5 depicts a network with contingent actions.

There two things to note here. First, actions can be made contingent on whether or not previous actions were executed. Consequently, it is straightforward to represent a contingent action sequence (i.e., a contingency plan). Second, there is no requirement that action selection be a deterministic function of the situation. It could be probabilistic, to reflect possible uncertainties about the agents ability to detect the true status of a situation. Alternatively, one could make the action contingent on a sensor report and make the sensor report a probabilistic function of the situation.

### 3.5 MULTIPLE LEVELS OF ABSTRACTION.

Many planners use operators at varying levels of abstraction. As a result, there may be plans that are only partially detailed. In order to build PE-nets for such plans, it is necessary to have probabilistic action models for operators at each level of abstraction. High level actions can be pasted onto the network in exactly the same manner as less abstract actions.

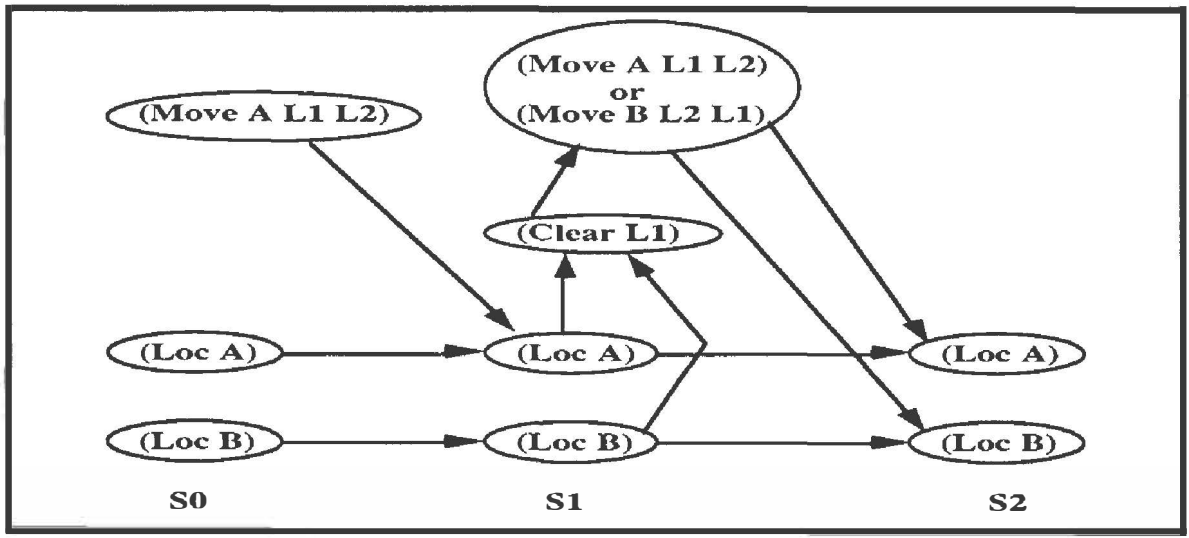

**Figure 5. PE-net for plan with contingent actions**

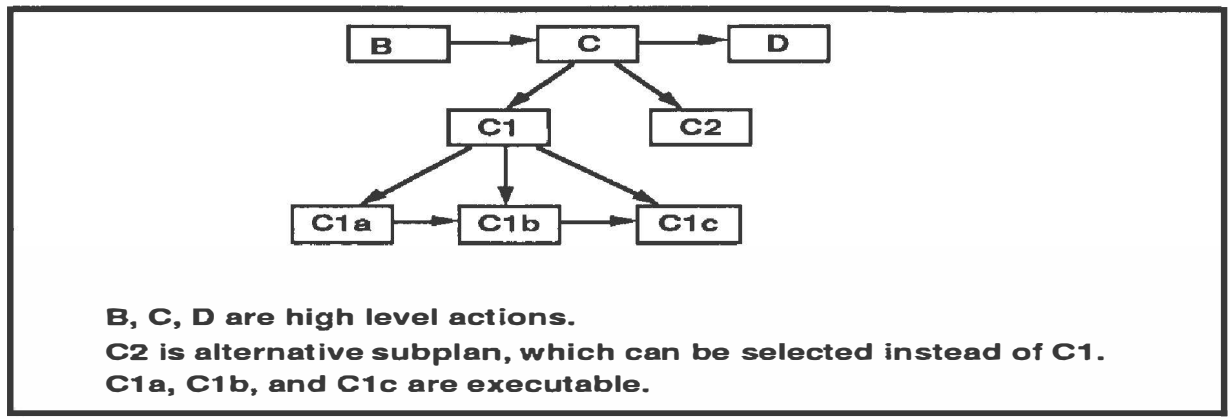

**B, C, D are high level actions.**
**C2 is alternative subplan, which can be selected instead of C1.**
**C1a, C1b, and C1c are executable.**

**Figure 6.    Example hierarchical plan.**

When an abstract operator is expanded, the PE-subnet for that expansion should be pasted onto the overall PE-net. There are two things to note about the PE-subnet. First, not only should it contain the actions that are selected to be part of the plan, but it should also contain the actions that were enumerated, but not selected. To illustrate, consider the plan in Figure 6. B, C and D are abstract actions, each capable of expansion. After expanding C, it turns out that there are two possible approaches to achieving C, namely C1 and C2. C1 is selected for inclusion in the plan and is further expanded to the sequence of actions C1a, C1b and C1c. To construct the PE-net, a subnet that combines C1 and C2 into a contingent action node is constructed and pasted onto the PE-net. This requires that the conditions be enumerate under which the alternative action will be selected. After this, the subnet for the C1a, C1b, C1c sequence is constructed and pasted onto the PE-net. The second thing to note is that when a PE subnet for an expanded subplan is pasted onto a PE net, it doesn't necessarily override everything in the more abstract action model. There may be consequence predicates of the higher level action model that are not mentioned in the lower level action models.

One advantage of using PE-nets to evaluate hierarchical plans is that the PE-net can be processed to estimate both the probability that the current plan will succeed and the probability that the current plan will lead to success (i.e., the probability that the plan can be successfully modified during execution). The probability that the current plan will succeed is the joint probability that the goal conditions (represented as specific states on specified predicates) will be true in the final situation and that the (most detailed) steps in the current plan will be executed, while the probability that the plan will lead to success is just the probability that the target conditions will be true in the final situation.

### 3.6    OVERLAPPING ACTIONS, DURING CONDITIONS AND EFFECTS.

In AP, a planning agent may plan the coordinated activity of multiple execution agents. Although the plan for each execution agent is linear, the overall plan will contain multiple simultaneous actions with interlocking start and end situations.    To relate the effects of overlapping actions, AP action models use *during conditions* and *during effects*. A during condition is a proposition that must be true during execution of an action in order for some effect to occur. Similarly, some effects occur during the execution of an action, rather than in the end situation of that action.

If the probabilistic action and persistence models do not mention specific times (see below), then PE-net construction for plans with overlapping actions proceeds by arbitrarily selecting a linear ordering on the situations that is consistent with the interlock constraints, and then pasting onto the PE-net any during conditions and effects of an action for the nodes in the situations between the start and end situation of that action. The probability estimates derived from such a PE-net have two useful characteristics. First, they are minimum estimates. This is because the planning agent can choose to further constrain the plan so that the execution agents will execute the actions in a way that satisfies the linear ordering on the situations.    Second, in practical applications the probability estimates of the goal conditions are not likely to change substantially if a different linear ordering is selected.    This is because nonlinear planners (such as AP) are specifically designed to impose order constraints whenever the current constraints leave attainment of the goal conditions in doubt. Consequently, while it is certainly possible for a nonlinear planner to miss an important order constraint, a planner that does this often is unlikely to transition to practical applications.

### 3.7    REFERENCES    TO    SPECIFIC    TIMES AND DURATIONS.

One can easily introduce time into situations by adding a predicate for clock time and having action models that assign a probability distribution over the clock time in the end situation conditioned on the clock time in the start situation.    If clock time are included, then the probabilistic persistence models can use time elapsed since the previous situation as a conditioning variable.

This approach works well for linear plans, where the sequence of situations are necessarily in temporal order no matter what the distribution of situation clock times. Unfortunately, this does not always hold for plans with overlapping actions. To illustrate the problem may result from overlapping actions , consider the plan in Figure 7a. In this plan actions A1 and A2 begin together. A1 takes either 2 or 4 minutes to complete, A2 takes 1 or 6 minutes. As a result, the clock time for $S_1$ is either 2 or 4, and for $S_2$ it is either 1 or 6. If the PE-net for this plan orders the situations $S_0$-->$S_1$-->$S_2$-->$S_3$, then there are possible states for Clock-time in $S_1$ that come after some states for Clock-time in $S_2$. As a result, the persistence models must condition the probability distribution over the other nodes in $S_1$ as a function of negative elapsed times. Obviously intolerable.

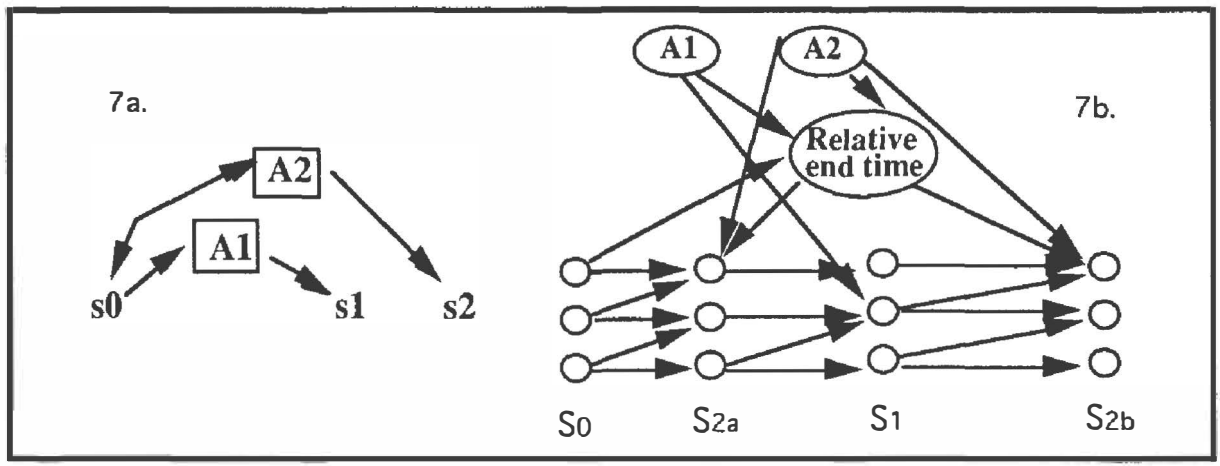

**Figure 7. Structure of PE-net for plan that lacks a clear temporal order on situations.**

A solution to this problem is to split situations so that the temporal ordering of the situations is guaranteed. For instance, as shown in Figure 7b, $S_2$ can be split into $S_{2a}$ and $S_{2b}$. A new node, Relative-end-time, is added. The probabilistic action model for A2 is pasted onto $S_{2a}$ whenever Relative-end-time is negative. Otherwise it is pasted onto $S_{2b}$. This solution guarantees that the situations are in temporal order, even though the clock times for the situations may overlap.

## 4. DISCUSSION

Our work to date suggests that automated procedures can be developed for constructing PE-nets for plans that contain a variety of complicating features. Belief networks do seem to provide an adequate formal foundation for probabilistic evaluation of plans, and automated construction of these nets is feasible.

Clearly of great concern is computational complexity. Our work to date suggests that for linear plans the number of nodes in a PE-net grows linearly with the length of a plan. However, unless care is taken, the number of node states will increase exponentially with the length of the plan. This will occur whenever multiple node states are generated for each node state in a previous situation. This problem can be mitigated somewhat by defining an "OTHER" node state, which combines into a single node state a set of node states that seem to have little relevance to evaluating the plan. In general, if the action and persistence models are carefully engineered, then we anticipate that the number of node states will increase linearly with the length of a linear plan.

Nonlinear plans are more problematic. If relative end time nodes are inserted then, as the example in Section 3.7 indicates, the number of situations will increase rapidly, where every situation will contain most of the primitive predicates mentioned in any of the action models. The rate of increase is not exponential, but it is substantial. Finally, exact processing of a belief net increases exponentially with the size of the network (Cooper, 1990). This suggests that approximate (e.g., monte carlo) algorithms should be used to process large PE-nets.